\pdfoutput=1

\documentclass[11pt]{article}

\usepackage{EMNLP2022}

\usepackage{times}
\usepackage{latexsym}
\usepackage{tabularx}
\usepackage{booktabs}
\usepackage{multirow}

\usepackage[T1]{fontenc}

\usepackage[utf8]{inputenc}

\usepackage{microtype}

\usepackage{inconsolata}

\usepackage{graphicx}
\usepackage{subcaption}

\usepackage[normalem]{ulem}

\title{Prompt-Tuning Can Be Much Better Than Fine-Tuning on Cross-lingual Understanding With Multilingual Language Models}

\author{Lifu Tu \and Caiming Xiong \and Yingbo Zhou \\
  Salesforce AI Research\\
  \texttt{\{ltu,cxiong,yingbo.zhou\}@salesforce.com} \\}

\begin{document}
\maketitle
\begin{abstract}
Pre-trained multilingual language models show significant performance gains for zero-shot cross-lingual
model transfer on a wide range of natural language understanding (NLU) tasks. Previously, for zero-shot cross-lingual evaluation, pre-trained models are only fine-tuned on English data and tested on a variety of target languages. In this paper, we do cross-lingual evaluation on various NLU tasks (sentence classification, sequence labeling, question answering) using prompt-tuning and compare it with fine-tuning. The results show that prompt tuning achieves much better cross-lingual transfer than fine-tuning across datasets, with only 0.1\% to 0.3\% tuned parameters. Additionally, we demonstrate through the analysis that prompt tuning can have better cross-lingual transferability of representations on downstream tasks with better aligned decision boundaries\footnote{Code 
is available at \url{https://github.com/salesforce/MPT}}.
\end{abstract}

\section{Introduction}

Large Multilingual language models~\citep{pires-etal-2019-multilingual,wu-dredze-2019-beto,conneau-etal-2020-unsupervised} show surprisingly impressive zero-shot cross-lingual transfer on NLP tasks, even though they are only trained from monolingual corpora. Recently, large-scale benchmarks such as XTREME~\citep{pmlr-v119-hu20b} and XGLUE~\citep{liang-etal-2020-xglue} are introduced for cross-lingual evaluation. 

In a cross-lingual transfer setting, models are only fine-tuned on the task-specific annotations in one language and evaluated in other languages. During fine-tuning, pre-trained language models are used for initialization and the entire model parameters are tuned on downstream tasks. While fine-tuning obtains strong performance, it is inefficient.
Also as shown in ~\citep{pmlr-v119-hu20b}, the cross-lingual transfer gap between the performance on the English test set and all other languages is large even with the best baseline XLM-R~\citep{conneau-etal-2020-unsupervised}.

Recently, prompt tuning, where only a small amount of additional parameters (i.e. prompts) is added and tuned, but the original model is kept frozen.
Much fewer parameters or no parameters
are tuned and thus the training is a lot more efficient.
Prompt tuning still performs worse than fine-tuning in lots of NLP tasks\citep{brown2020language,autoprompt:emnlp20,zhong-etal-2021-factual}. More recently, ~\citet{li-liang-2021-prefix,lester-etal-2021-power,hambardzumyan-etal-2021-warp} indicate prompt tuning is competitive with fine tuning on some of the NLU tasks. Language model capacity (e.g., 10 billion parameters) is a key ingredient for these approaches to succeed. More recently, ~\citep{liu-etal-2022-p} shows prompt tuning can also be comparable on several hard monolingual sequence
labeling tasks such as extractive question answers.

In this paper, we aim to investigate the effect of prompt tuning in cross-lingual tasks.We freeze the entire multilingual language model and tune task prompts on the English training set for downstream tasks (sentence classification, structure prediction, question answering). Even with medium size multilingual language model (less than 1 billion parameters), prompt tuning achieves much higher performance than fine-tuning on various NLU tasks.

According to the analysis results, prompt tuning does fewer changes to sentence representations than fine-tuning and keeps good cross-lingual sentence representations. We also find that the decision boundaries of different language sentence representations after prompt tuning on English data are almost aligned well. However, these decision boundaries of different languages after fine-tuning are a large difference. These aligned decision boundaries can lead to stronger cross-lingual transfer.

This work sheds light on the strong cross-lingual ability of prompt tuning. Our results suggest prompt tuning is better than fine-tuning on cross-lingual transfer.  
Our contributions are summarized as follows: we show that prompt tuning can perform much better as compared to fine-tuning for cross-lingual transfer; we also show prompt tuning works better in the case of the cross-lingual transfer due to the relative small robust changes it brings to the originally learned representations.

\section{Prompt-Tuning for Cross-Lingual Tasks}

\paragraph{Multilingual Language Models.} 

In the past years, lots of pre-trained multilingual language models come out: mBERT, XLM~\citep{NEURIPS2019_c04c19c2}, XLM-R~\citep{conneau-etal-2020-unsupervised}, 
etc. XLM-R~\citep{conneau-etal-2020-unsupervised}  significantly outperforms multilingual BERT (mBERT;~\citealp{devlin-etal-2019-bert}) on a variety of cross-lingual benchmarks XTREME~\citep{pmlr-v119-hu20b}. In some previous work~\citep{luo-etal-2021-veco,zhang-etal-2019-ernie}, XLM-R is also used for initialization to do another round of pretraining with parallel data to get the stronger cross-lingual ability. Previously, in the cross-lingual evaluation,  models are fine-tuned on the English training data but evaluated on all target languages. As far as we know, we are the first to explore prompt tuning on several hard multilingual NLP tasks including structure prediction and question answering

\begin{figure}[h]
    \centering
    \includegraphics[width=0.48\textwidth]{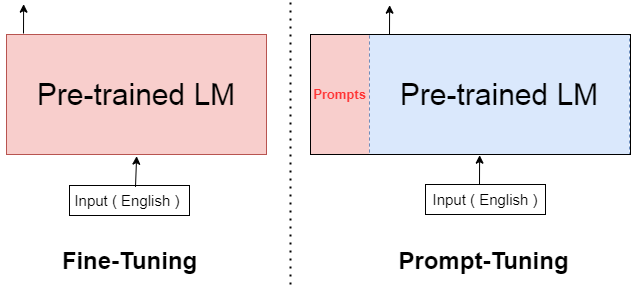}
    \caption{Two different approaches for cross-lingual evaluation when using large multilingual language model. \textbf{Left:} In fine-tuning, all model parameters are tuned on English task data. This setting is used in cross-lingual evaluation before. \textbf{Right:} In prompt tuning, only small ratio parameters are tuned. We use prefix prompts and use layer prompts in our experiments. }
    \label{fig:framework}
\end{figure}

\paragraph{Prompt Tuning.} Fine-tuning on large pre-trained language models leads to strong performance on downstream tasks, however, it is memory-consuming and lots of parameters need to save for each task. In prompt tuning, only a small part of the parameters ( e.g., prompts or task classifier ) are tuned during learning. However, it usually performs not as good as compared to fine-tuning. Recently, \citet{lester-etal-2021-power} find prompt tuning can be better than fine-tuning when the model size is not extremely large (10 billion parameters). Prefix-tuning~\citep{li-liang-2021-prefix} obtains
comparable performance for natural language generation tasks. \citet{liu-etal-2022-p} shows prompt tuning can be matched to fine-tuning on language understanding tasks even at hard sequence tagging tasks.   

We investigate prompt tuning on cross-lingual understanding on a pre-trained multilingual language model. The framework is shown in Figure~\ref{fig:framework}. Our setting is similar to ~\citet{li-liang-2021-prefix,liu-etal-2022-p}. The continuous prompts are added as prefix tokens and tuned during learning. In the implementation, the prompts are operated as past keys and values in each transformer layer. Each transformer layer has separated prompts. These continuous prompts are optimized, but multilingual language model parameters are frozen.

\section{Experiments Setup}

\begin{table*}[t!]
\centering
\scalebox{0.78}{\begin{tabular}{lccccccc}
\toprule
\multirow{2}{*}{\bf Model} & \multicolumn{2}{c}{\bf Sentence Classification} & \multicolumn{1}{c}{\bf Structured Prediction} & \multicolumn{3}{c}{\bf Question Answering}   \\
& XNLI & PAWS-X & UD-POS  & XQuAD  & TyDiQA  & \\ \midrule
Metrics & Acc. & Acc. & F1  & F1 / EM  & F1 / EM  \\ 
\midrule
\multicolumn{2}{l}{\textbf{Fine Tuning}} \\
\midrule
\textsc{mBERT}* & 65.4 & 81.9 & 70.3 & 64.5 / 49.4   & 59.7 / 43.9   \\
\textsc{XLM-R-Large}* & \underline{79.2} & 86.4 &  72.6 &  76.6 / 60.8 & 65.1 / 45.0   \\
\textsc{XLM-R-Large}$^{+}$ & \underline{79.2} & - &  \underline{75.0} &  77.2 / 61.6  &  64.3 / 45.8   \\
\textsc{XLM-R-Large (our)} & 78.8 (0.2) & \underline{87.9} (0.5) & 74.4 (0.7)  &  \underline{77.3} (0.4) / \underline{61.8} (0.5)   &   \underline{70.1} (0.6) / \underline{51.7} (2.7)   \\
\midrule
\multicolumn{2}{l}{\textbf{Prompt Tuning}} \\
\midrule
\textsc{XLM-R-Large} & \textbf{79.9} (0.1) & \textbf{88.4} (0.3) &  \textbf{75.4} (0.2) &  \textbf{79.0} (0.2) / \textbf{64.1} (0.4)  & \textbf{71.5} (0.4) / \textbf{55.1} (0.6) \\
\bottomrule
\end{tabular}}
\caption{Zero-shot cross-lingual transfer evaluation results (with standard deviation) on XTREME structured prediction, question answering, and sentence classification tasks. For both fine tuning and prompt tuning, models are only fine-tuned on the English training data but evaluated on all target languages. Baseline fine-tuning results with ``*'' and ``+'' are taken from~\citep{pmlr-v119-hu20b} and ~\citep{ruder-etal-2021-xtreme} respectively. More results are shown in the Appendix.}
\vspace{-0.5cm}
\label{table:overview}
\end{table*}

\subsection{Datasets.}
We perform experiments on four datasets included in XTREME: cross-lingual natural language
inference (XNLI; ~\citealp{conneau-etal-2018-xnli}), cross-lingual adversarial dataset for paraphrase identification (PAWS-X; ~\citealp{yang-etal-2019-paws}), part-of-speech tagging on the Universal Dependencies (UD-POS; ~\citealp{nivre2018universal}), cross-lingual question answering on XQuAD~\citep{artetxe-etal-2020-cross} and TyDiQA-GoldP~\citep{clark-etal-2020-tydi}. Three categories of downstream tasks are included: (1) sentence classification); (2) structure prediction; (3) question answering.

\subsection{Training Details.}
Our frozen models are built on the top of the pretrained XLM-R checkpoint of LARGE size with about 560M parameters. Previous work~\citep{pmlr-v119-hu20b} shows it achieves stronger performance than mBERT\footnote{Some preliminary results are obtained with mBERT.}. All our experiments were run with Huggingface~\citep{wolf-etal-2020-transformers}. More details are in the appendix. 

\paragraph{Prompt Length.} Prompt length usually plays an important role in prompt tuning. In our experiments, we treat this as a hyper-parameter. Longer prompt length often leads 
to have higher performance. In our experiments, prompt length is set to 16 or 32 and tuned on the English validation set.

\section{Results}

\paragraph{Tuned Parameter Sizes Comparison}
For the prompt tuning test results in Table~\ref{table:overview}, we
did limited tuning on prompt length. The prompt length is 16, except prompt length for task XNLI is 32. With only 0.1\% to 0.3\% additional prompt parameters as compared to the original model, the framework already 
demonstrates strong cross-lingual results.  

\paragraph{Overall Results}
Table~\ref{table:overview} shows the zero-shot cross-lingual results on four different tasks. Prompt tuning performs much better than fine-tuning, especially for hard sequence task question answering. And prompt tuning is also with smaller variance.

Previously, although with parallel data or more monolingual data, cross-lingual transfer results~\citep{zhang-etal-2019-ernie,luo-etal-2021-veco,ruder-etal-2021-xtreme} on question answering and structured prediction tasks improved only slightly. With prompt tuning, there is larger performance gains for question answering and structured prediction tasks. It suggests that prompt tuning is a better tuning method for cross-lingual transfer.

\paragraph{Cross-lingual Transfer Gap} According to the above result, on average, prompt tuning achieves better performance than fine tuning.  Table~\ref{table:gap} shows the cross-lingual transfer gap of the two different tuning methods. Prompt tuning can also reduce the gap significantly. 
\begin{table}[ht!]
\centering
\small
\scalebox{0.9}{
\begin{tabular}{c|c|c|c|c|c|}

 \multicolumn{1}{l}{}  & XNLI & PAWS-X & UD-POS & XQuAD \\ 
\midrule
 Fine Tuning & 10.2 & 12.4 & 24.3 & 16.3  \\
\midrule 
 Prompt Tuning & 9.7 &  8.7 & 20.7 & 14.5   \\ 
\hline
\end{tabular}}
\caption{Cross-lingual transfer gap of the two tuning methods. The cross-lingual transfer gap is the performance difference
between English test set and the average of the other languages. The smaller is better.}
\label{table:gap}
\end{table}

\paragraph{Discussion} In our preliminary experiments, for the smaller size model (e.g., mBERT), prompt tuning perform a little worse than fine tuning on English, and match the performance of fine-tuning on all languages. The language model size still matter. There is still some space for smaller size model. This also indicates potential for future work with better prompt tuning method.

\section{Analysis}
In order to perform some analysis on prompt tuning and fine tuning, we select 1000 samples for each language (en, de, es, fr, ja, ko, zh ) from PAWS-X~\citep{yang-etal-2019-paws} dataset. For each English language sample in our selections, there is a human translated sample from the other six languages. \footnote{Each sample in PAWS-X dataset is a sentence pair. In the following experiments, we treat the representations at CLS token as the sample sentence representations.} 

\begin{figure*}[h!]
  \centering
  \vspace{-0.5cm}
  \begin{subfigure}[b]{0.49\linewidth}
    \includegraphics[width=\linewidth]{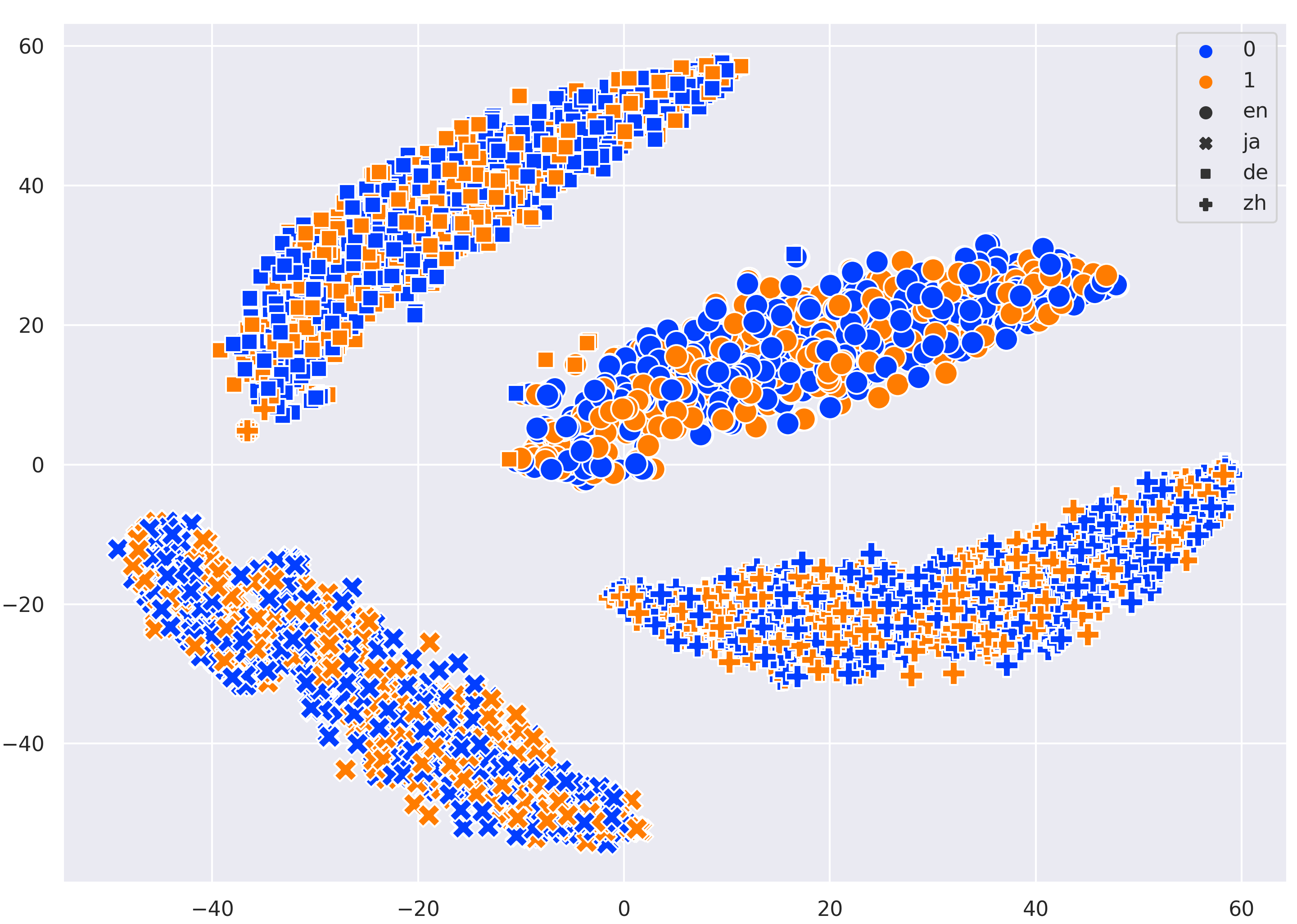}
    \caption{Visualization before fine tuning (FT). }
  \end{subfigure}
    \begin{subfigure}[b]{0.49\linewidth}
    \includegraphics[width=\linewidth]{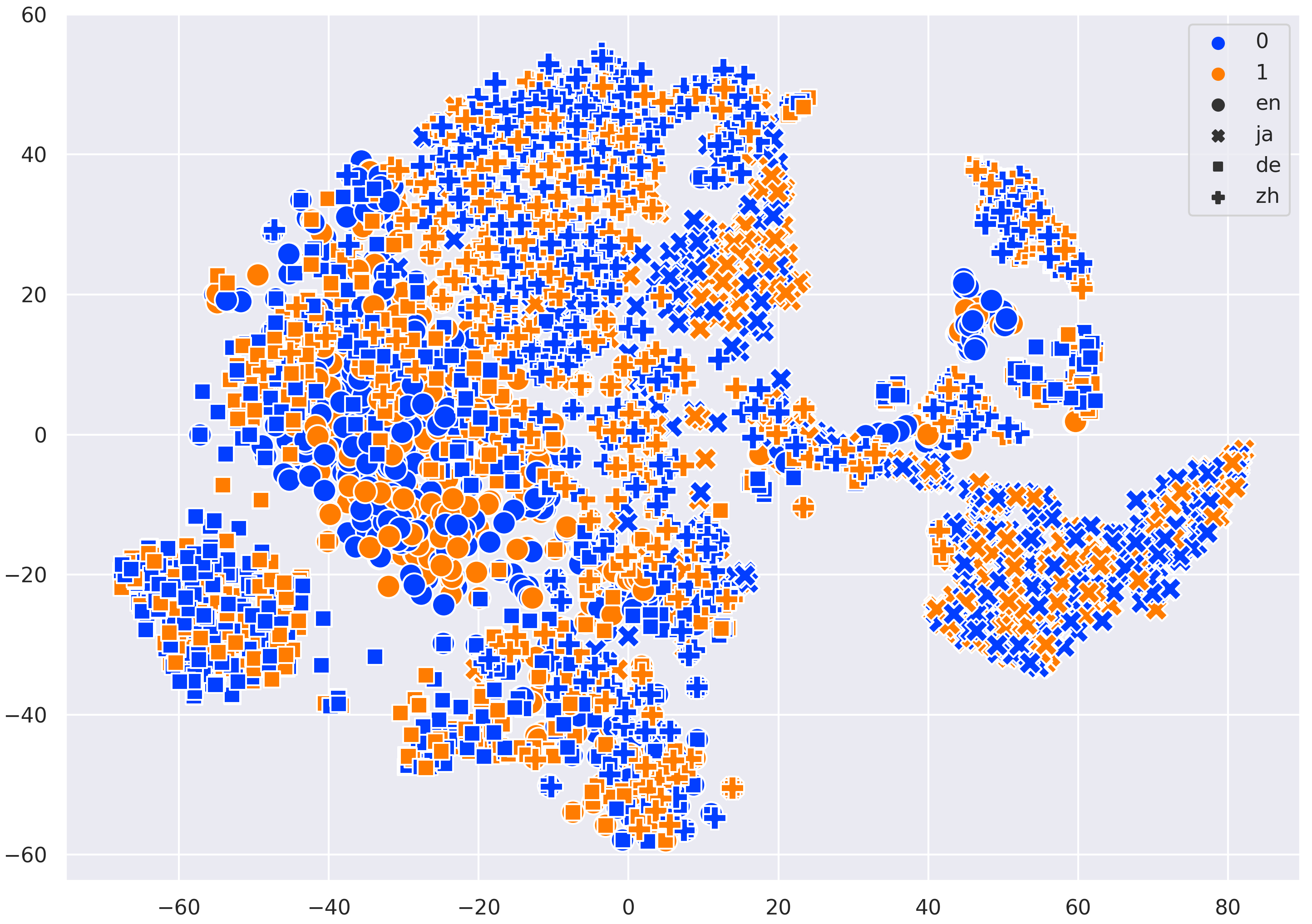}
    \caption{Visualization before prompt tuning (PT).}
  \end{subfigure}
  \begin{subfigure}[b]{0.49\linewidth}
    \includegraphics[width=\linewidth]{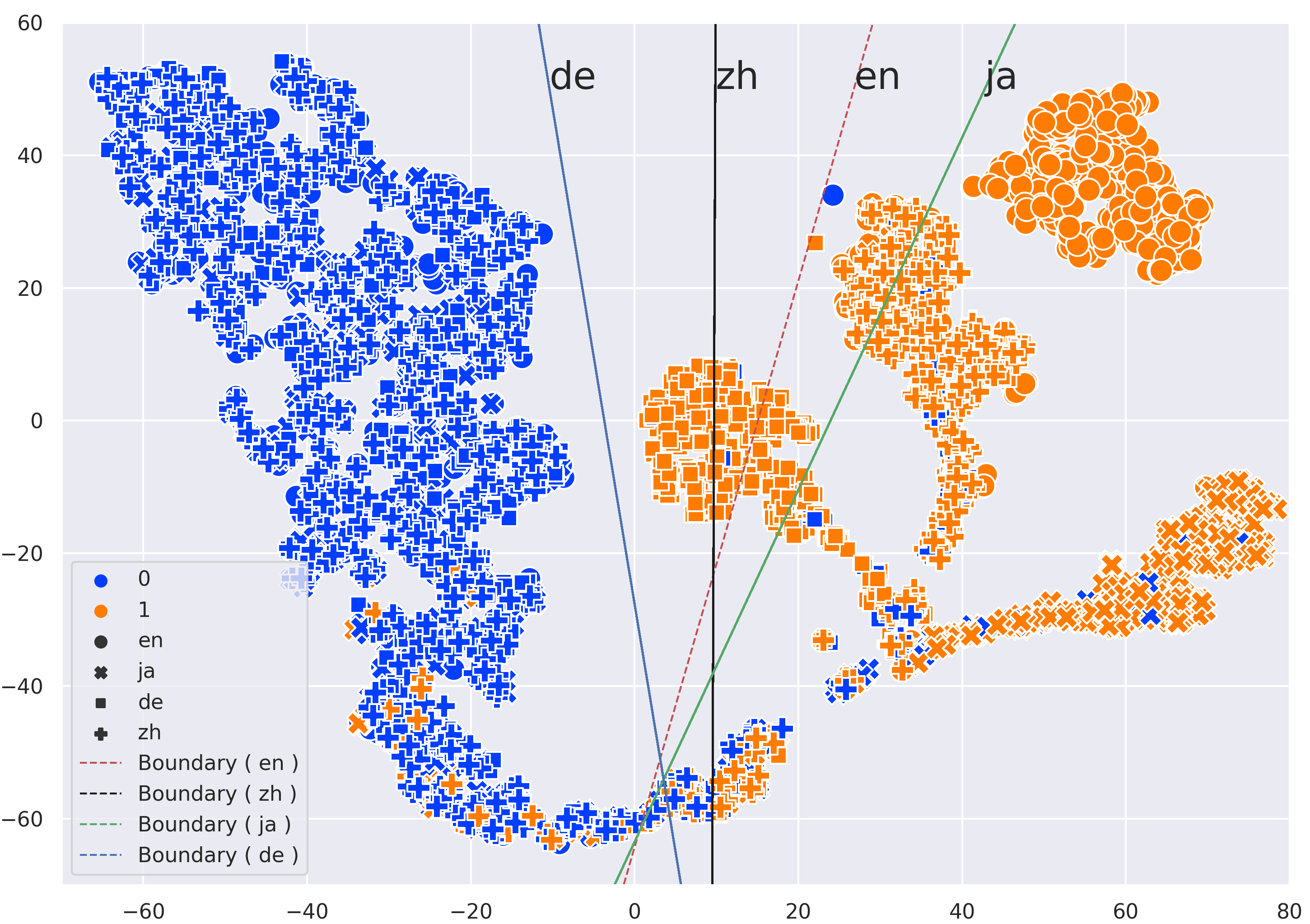}
    \caption{Decision boundaries after fine tuning (FT). }
  \end{subfigure}
    \begin{subfigure}[b]{0.49\linewidth}
    \includegraphics[width=\linewidth]{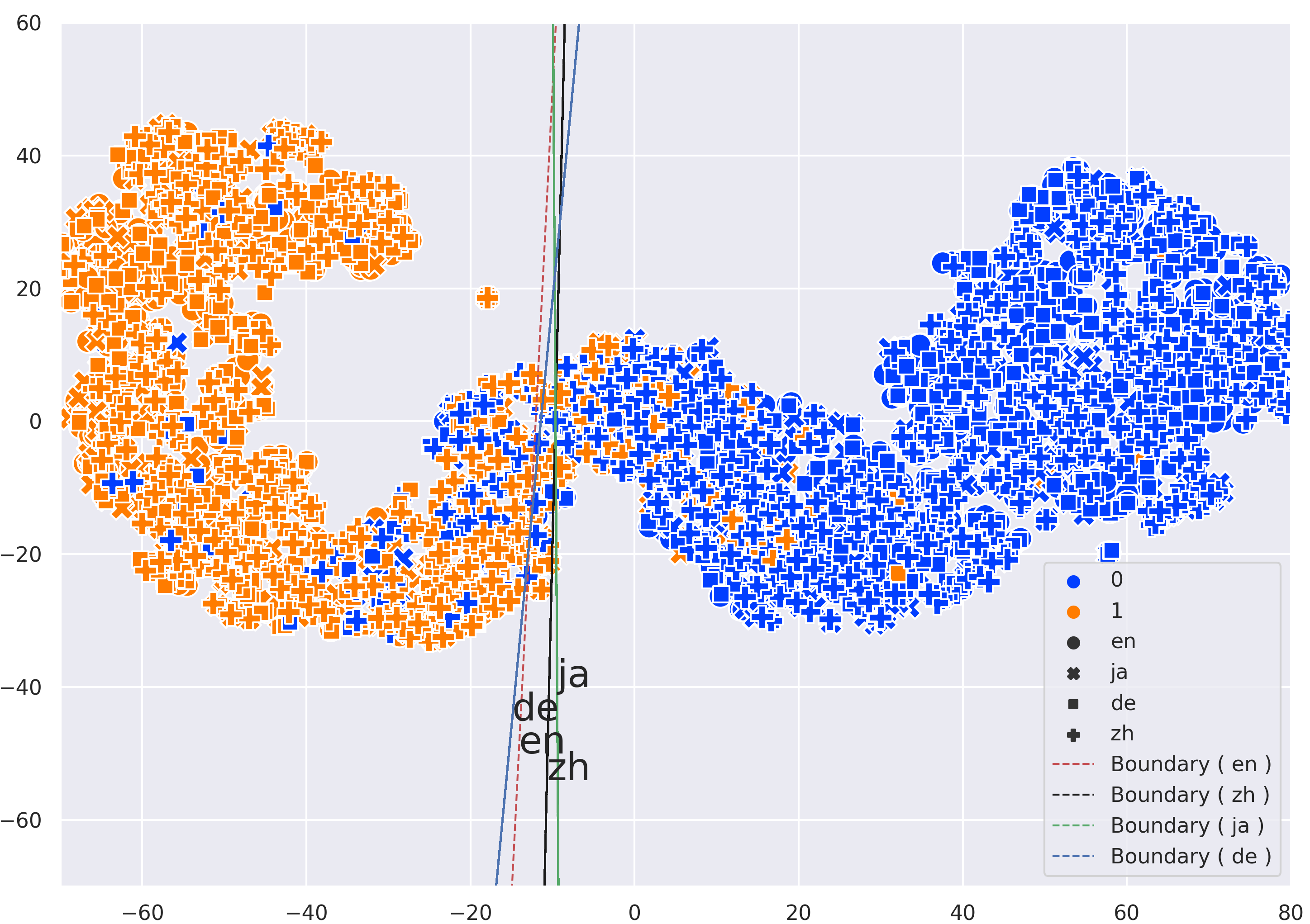}
    \caption{Decision boundaries after prompt tuning (PT).}
  \end{subfigure}
  \caption{T-SNE visualization of representations of four languages (en: English; de: German; ja: Japanese; zh: Chinese) 
 before and after two different tuning methods on English task data. The decision boundaries after prompt tuning is aligned much better. 
  \label{fig:decision}
  }

\end{figure*}

Figure~\ref{fig:decision} shows t-SNE visualization of sample representations from frozen multilingual language model XLM-R. Samples'  representations are clustered well respect to languages, however, there is weak correlation with labels.

\subsection{Language Representation Changes}

For each tuning method (fine-tuning and prompt-tuning), Table~\ref{table:change_result} shows the cosine similarity of representations from frozen language model and tuned model. According to the results, both of two tuned method make notable change on sentence representations. However, the average cosine similarity of fine-tuning is much smaller. It indicates that fine-tuning leads much larger changes on sentence representations than prompt tuning. We can also see representation changes is larger when tuning is on MNLI, while prompt tuning still has less changes on representations.

\begin{table}[ht!]
\centering
\small
\scalebox{0.9}{
\begin{tabular}{c|c|c|c|c|c|c|c|c|}

 \multicolumn{1}{l}{}  & en & de & es & fr & ja & ko & zh \\ 
\midrule
\multicolumn{4}{l}{\textbf{Training on PAWS}} \\
\midrule
 FT & 25.2 & 26.5 & 24.5 & 25.2 & 18.9 & 15.0 & 22.6 \\
\midrule
 PT & 57.6 & 56.8 & 57.2 & 57.7 & 58.7 & 59.4 & 59.5\\ 

\midrule
\multicolumn{4}{l}{\textbf{Training on MNLI}} \\
\midrule
 FT & -16.9 & -19.1 & -16.3 & -14.5 & -16.7 & -11.8 & -14.9 \\
\midrule 
 PT & 32.2 &  32.1 & 31.2 & 32.1 & 33.8 & 36.0 & 35.8\\ 
\hline
\end{tabular}}
\caption{Cosine similarity (\%) of representations after tuning for each language. FT: fine-tuning; PT: prompt tuning. These checkpoints are on tuned on two English datasets: PAWS and MNLI.\footnote{They are uses as training sets when evaluating XNLI and PAWS-X separately.}}
\label{table:change_result}
\end{table}

\subsection{Cross-lingual Alignment After Tuning}


We compute the averaged cosine similarity of all the 1000 translation pairs for each language pair <en , xx>, where xx is de, es, fr, ja, ko or zh. We also compute averaged cosine similarity of all the 1000*999/2 non-translations for each language pair. 
As shown in Table~\ref{table:change_result}, both fine tuning and prompt tuning are doing well. Prompt tuning has the advantage in the sense that they change the representation more mildly, still have high cosine similarity on translation pairs. This resulted in more robust transfer and less overfitting. 

\begin{table}[t!]
\centering
\small
\scalebox{0.9}{\begin{tabular}{c|c|c|c|c|c|c|c|}
 \multicolumn{1}{l|}{}  & en-de &  en-es & en-fr & en-ja & en-ko & en-zh \\ 

 \midrule
\multicolumn{4}{l}{\textbf{Training on MNLI}} \\
\midrule
 FT & 81.5 & 85.4 & 83.0 & 71.8 & 68.2 & 73.9   \\
 FT-neg & 52.6 & 53.1 & 52.8 & 51.5 & 50.6 &  50.0   \\
 rel-diff (\%) & 54.8 & 60.8 & 57.2 & 39.4 & 34.8 &  47.8   \\
\hline 
 PT & 96.4 & 97.3 & 96.6 & 94.8 & 93.8 &  95.0\\ 
 PT-neg & 91.0  & 91.1 & 90.8 & 90.5  & 90.1 & 90.2    \\
 rel-diff (\%) & 5.9 & 6.8 & 6.4 & 4.8 & 4.1 &  5.3   \\

\midrule
\multicolumn{4}{l}{\textbf{Training on PAWS}} \\
 FT & 90.4 & 92.1 & 88.8 & 76.8 & 75.3 & 82.0    \\
 FT-neg & 13.3 & 13.2 & 13.4 & 14.3 & 14.4 & 13.6     \\
 rel-diff (\%) & 580  & 598 & 563 & 437 & 423 & 503    \\

\midrule
 PT & 98.4 & 98.6 & 98.3 & 96.3 & 96.0 & 96.7 \\ 
 PT-neg & 88.1  & 88.1 & 88.3  & 89.1  & 89.4 & 88.9\\
 rel-diff (\%) & 11.7 & 11.9 & 11.3 & 8.1 & 7.4 & 8.8    \\
\hline
\end{tabular}}
\caption{Cosine similarity (\%) of translation pairs after tuning on two English dataset: MNLI and PAWS. ``-neg`` means the average cosine similarity of non-translations for each language pair. ``rel-diff`` means the relative difference between translation and non-translations. Two different tuning method are shown, one is fine-tuning (FT), the other is prompt tuning (PT). }
\label{table:align_result}
\end{table}

\subsection{Decision Boundaries}

Prompt tuning keeps high cross-lingual alignment with fewer changes in the previous subsections. The general level of the learned representations' quality is still unknown, though. The learned representations quality are examined in this subsection.

Figure~\ref{fig:decision} (a) and (b) show t-SNE visualization of representations before two different tuning methods. Each dot in the two figures is a PAWS-X sample from four languages: German (de), zh (Chinese), en (English), ja (Japanese). The blue sample is a paraphrase, the orange sample is a non-paraphrase. Samples of the same language are grouped together. However, label information is missing from sample representations.

Figure~\ref{fig:decision} (c) and (d) shows t-SNE~\citep{JMLR:v9:vandermaaten08a} visualization after fine tuning (FT) and prompt tuning (PT). After tuning, both have reasonable and nice separated representations. For each language, we also plot logistic regression decision boundary for these t-SNE embeddings. The decision boundaries for various languages vary significantly after fine tuning. The English decision boundary can not separate well on German samples. After prompt tuning, the decision boundaries of the four languages are surprisingly aligned well. This suggest that prompt tuning learns better language-independent classifier than fine tuning, although the tuning is only on English training set.

\section{Related Work}

Recently, several previous works show prompt tuning for multilingual language models. ~\citet{winata-etal-2021-language} shows the multilingual skills of large pre-trained models with few examples. ~\citet{zhao-schutze-2021-discrete,Huang2022ZeroshotCT,qi-etal-2022-enhancing} shows new proposed prompt tuning methods. The goal of our work is different from theirs. We show prompt tuning is better than fine-tuning for cross-lingual evaluation. We have a conclusion that our prompt tuning achieves higher performance than fine-tuning consistently in the setting. 

Previous work~\citep{zhao-schutze-2021-discrete,Huang2022ZeroshotCT,qi-etal-2022-enhancing} only experimented on the sentence classification task. Hard sequence tagging tasks and question answering is not explored or the settings are in low resource regimes. We investigate cross-lingual transfer ability on various NLU tasks from XTREME~\citep{pmlr-v119-hu20b}, which is one of the important cross-lingual transfer evaluation benchmarks. Sentence classification, sequence labeling, and question answering are included.

\section{Conclusion}
In this work, we compared prompt tuning and fine tuning on cross-lingual understanding with multilingual languages models, finding that prompt tuning achieves a better performance. This suggest that it is promissing to use prompt tuning on cross-lingual transfer.

\section*{Limitations}

In this work, we investigate the effects of prompt tuning on cross-lingual understanding and empirically demonstrate some promising outcomes. We need a lot of GPU resources to complete our experiments. The experiments on large size pretrained multilingual language models are conducted on A100s with 40G memory. Training can be accelerated by using large batches. 

 This is a preliminary exploration of prompt tuning on cross-lingual transfer. In this work, encoder-only models are explored on natural language understanding tasks in the paper. Future work may also involve encoder-decoder models and other tasks.

\section*{Acknowledgements}
We would like to thank Salesforce AI Research team for helpful discussions, and the reviewers for insightful comments.

\bibliography{anthology,custom}

\begin{thebibliography}{29}
\expandafter\ifx\csname natexlab\endcsname\relax\def\natexlab#1{#1}\fi

\bibitem[{Artetxe et~al.(2020)Artetxe, Ruder, and
  Yogatama}]{artetxe-etal-2020-cross}
Mikel Artetxe, Sebastian Ruder, and Dani Yogatama. 2020.
\newblock \href {https://doi.org/10.18653/v1/2020.acl-main.421} {On the
  cross-lingual transferability of monolingual representations}.
\newblock In \emph{Proceedings of the 58th Annual Meeting of the Association
  for Computational Linguistics}, pages 4623--4637, Online. Association for
  Computational Linguistics.

\bibitem[{Brown et~al.(2020)Brown, Mann, Ryder, Subbiah, Kaplan, Dhariwal,
  Neelakantan, Shyam, Sastry, Askell et~al.}]{brown2020language}
Tom~B Brown, Benjamin Mann, Nick Ryder, Melanie Subbiah, Jared Kaplan, Prafulla
  Dhariwal, Arvind Neelakantan, Pranav Shyam, Girish Sastry, Amanda Askell,
  et~al. 2020.
\newblock Language models are few-shot learners.
\newblock \emph{arXiv preprint arXiv:2005.14165}.

\bibitem[{Clark et~al.(2020)Clark, Choi, Collins, Garrette, Kwiatkowski,
  Nikolaev, and Palomaki}]{clark-etal-2020-tydi}
Jonathan~H. Clark, Eunsol Choi, Michael Collins, Dan Garrette, Tom Kwiatkowski,
  Vitaly Nikolaev, and Jennimaria Palomaki. 2020.
\newblock \href {https://doi.org/10.1162/tacl_a_00317} {{T}y{D}i {QA}: A
  benchmark for information-seeking question answering in typologically diverse
  languages}.
\newblock \emph{Transactions of the Association for Computational Linguistics},
  8:454--470.

\bibitem[{Conneau et~al.(2020)Conneau, Khandelwal, Goyal, Chaudhary, Wenzek,
  Guzm{\'a}n, Grave, Ott, Zettlemoyer, and
  Stoyanov}]{conneau-etal-2020-unsupervised}
Alexis Conneau, Kartikay Khandelwal, Naman Goyal, Vishrav Chaudhary, Guillaume
  Wenzek, Francisco Guzm{\'a}n, Edouard Grave, Myle Ott, Luke Zettlemoyer, and
  Veselin Stoyanov. 2020.
\newblock \href {https://doi.org/10.18653/v1/2020.acl-main.747} {Unsupervised
  cross-lingual representation learning at scale}.
\newblock In \emph{Proceedings of the 58th Annual Meeting of the Association
  for Computational Linguistics}, pages 8440--8451, Online. Association for
  Computational Linguistics.

\bibitem[{CONNEAU and Lample(2019)}]{NEURIPS2019_c04c19c2}
Alexis CONNEAU and Guillaume Lample. 2019.
\newblock \href
  {https://proceedings.neurips.cc/paper/2019/file/c04c19c2c2474dbf5f7ac4372c5b9af1-Paper.pdf}
  {Cross-lingual language model pretraining}.
\newblock In \emph{Advances in Neural Information Processing Systems},
  volume~32. Curran Associates, Inc.

\bibitem[{Conneau et~al.(2018)Conneau, Rinott, Lample, Williams, Bowman,
  Schwenk, and Stoyanov}]{conneau-etal-2018-xnli}
Alexis Conneau, Ruty Rinott, Guillaume Lample, Adina Williams, Samuel Bowman,
  Holger Schwenk, and Veselin Stoyanov. 2018.
\newblock \href {https://doi.org/10.18653/v1/D18-1269} {{XNLI}: Evaluating
  cross-lingual sentence representations}.
\newblock In \emph{Proceedings of the 2018 Conference on Empirical Methods in
  Natural Language Processing}, pages 2475--2485, Brussels, Belgium.
  Association for Computational Linguistics.

\bibitem[{Devlin et~al.(2019)Devlin, Chang, Lee, and
  Toutanova}]{devlin-etal-2019-bert}
Jacob Devlin, Ming-Wei Chang, Kenton Lee, and Kristina Toutanova. 2019.
\newblock \href {https://doi.org/10.18653/v1/N19-1423} {{BERT}: Pre-training of
  deep bidirectional transformers for language understanding}.
\newblock In \emph{Proceedings of the 2019 Conference of the North {A}merican
  Chapter of the Association for Computational Linguistics: Human Language
  Technologies, Volume 1 (Long and Short Papers)}, pages 4171--4186,
  Minneapolis, Minnesota. Association for Computational Linguistics.

\bibitem[{Hambardzumyan et~al.(2021)Hambardzumyan, Khachatrian, and
  May}]{hambardzumyan-etal-2021-warp}
Karen Hambardzumyan, Hrant Khachatrian, and Jonathan May. 2021.
\newblock \href {https://doi.org/10.18653/v1/2021.acl-long.381} {{WARP}:
  {W}ord-level {A}dversarial {R}e{P}rogramming}.
\newblock In \emph{Proceedings of the 59th Annual Meeting of the Association
  for Computational Linguistics and the 11th International Joint Conference on
  Natural Language Processing (Volume 1: Long Papers)}, pages 4921--4933,
  Online. Association for Computational Linguistics.

\bibitem[{Hu et~al.(2020)Hu, Ruder, Siddhant, Neubig, Firat, and
  Johnson}]{pmlr-v119-hu20b}
Junjie Hu, Sebastian Ruder, Aditya Siddhant, Graham Neubig, Orhan Firat, and
  Melvin Johnson. 2020.
\newblock \href {https://proceedings.mlr.press/v119/hu20b.html} {{XTREME}: A
  massively multilingual multi-task benchmark for evaluating cross-lingual
  generalisation}.
\newblock In \emph{Proceedings of the 37th International Conference on Machine
  Learning}, volume 119 of \emph{Proceedings of Machine Learning Research},
  pages 4411--4421. PMLR.

\bibitem[{Huang et~al.(2022)Huang, Ma, Zhang, Wei, and
  Wang}]{Huang2022ZeroshotCT}
Lianzhe Huang, Shuming Ma, Dongdong Zhang, Furu Wei, and Houfeng Wang. 2022.
\newblock Zero-shot cross-lingual transfer of prompt-based tuning with a
  unified multilingual prompt.
\newblock \emph{ArXiv}, abs/2202.11451.

\bibitem[{Kingma and Ba(2015)}]{kingma2015adam}
Diederik~P. Kingma and Jimmy Ba. 2015.
\newblock \href {http://arxiv.org/abs/1412.6980} {Adam: {A} method for
  stochastic optimization}.
\newblock In \emph{3rd International Conference on Learning Representations,
  {ICLR} 2015, San Diego, CA, USA, May 7-9, 2015, Conference Track
  Proceedings}.

\bibitem[{Lester et~al.(2021)Lester, Al-Rfou, and
  Constant}]{lester-etal-2021-power}
Brian Lester, Rami Al-Rfou, and Noah Constant. 2021.
\newblock \href {https://doi.org/10.18653/v1/2021.emnlp-main.243} {The power of
  scale for parameter-efficient prompt tuning}.
\newblock In \emph{Proceedings of the 2021 Conference on Empirical Methods in
  Natural Language Processing}, pages 3045--3059, Online and Punta Cana,
  Dominican Republic. Association for Computational Linguistics.

\bibitem[{Li and Liang(2021)}]{li-liang-2021-prefix}
Xiang~Lisa Li and Percy Liang. 2021.
\newblock \href {https://doi.org/10.18653/v1/2021.acl-long.353} {Prefix-tuning:
  Optimizing continuous prompts for generation}.
\newblock In \emph{Proceedings of the 59th Annual Meeting of the Association
  for Computational Linguistics and the 11th International Joint Conference on
  Natural Language Processing (Volume 1: Long Papers)}, pages 4582--4597,
  Online. Association for Computational Linguistics.

\bibitem[{Liang et~al.(2020)Liang, Duan, Gong, Wu, Guo, Qi, Gong, Shou, Jiang,
  Cao, Fan, Zhang, Agrawal, Cui, Wei, Bharti, Qiao, Chen, Wu, Liu, Yang,
  Campos, Majumder, and Zhou}]{liang-etal-2020-xglue}
Yaobo Liang, Nan Duan, Yeyun Gong, Ning Wu, Fenfei Guo, Weizhen Qi, Ming Gong,
  Linjun Shou, Daxin Jiang, Guihong Cao, Xiaodong Fan, Ruofei Zhang, Rahul
  Agrawal, Edward Cui, Sining Wei, Taroon Bharti, Ying Qiao, Jiun-Hung Chen,
  Winnie Wu, Shuguang Liu, Fan Yang, Daniel Campos, Rangan Majumder, and Ming
  Zhou. 2020.
\newblock \href {https://doi.org/10.18653/v1/2020.emnlp-main.484} {{XGLUE}: A
  new benchmark datasetfor cross-lingual pre-training, understanding and
  generation}.
\newblock In \emph{Proceedings of the 2020 Conference on Empirical Methods in
  Natural Language Processing (EMNLP)}, pages 6008--6018, Online. Association
  for Computational Linguistics.

\bibitem[{Liu et~al.(2022)Liu, Ji, Fu, Tam, Du, Yang, and
  Tang}]{liu-etal-2022-p}
Xiao Liu, Kaixuan Ji, Yicheng Fu, Weng Tam, Zhengxiao Du, Zhilin Yang, and Jie
  Tang. 2022.
\newblock \href {https://doi.org/10.18653/v1/2022.acl-short.8} {{P}-tuning:
  Prompt tuning can be comparable to fine-tuning across scales and tasks}.
\newblock In \emph{Proceedings of the 60th Annual Meeting of the Association
  for Computational Linguistics (Volume 2: Short Papers)}, pages 61--68,
  Dublin, Ireland. Association for Computational Linguistics.

\bibitem[{Luo et~al.(2021)Luo, Wang, Liu, Liu, Bi, Huang, Huang, and
  Si}]{luo-etal-2021-veco}
Fuli Luo, Wei Wang, Jiahao Liu, Yijia Liu, Bin Bi, Songfang Huang, Fei Huang,
  and Luo Si. 2021.
\newblock \href {https://doi.org/10.18653/v1/2021.acl-long.308} {{VECO}:
  Variable and flexible cross-lingual pre-training for language understanding
  and generation}.
\newblock In \emph{Proceedings of the 59th Annual Meeting of the Association
  for Computational Linguistics and the 11th International Joint Conference on
  Natural Language Processing (Volume 1: Long Papers)}, pages 3980--3994,
  Online. Association for Computational Linguistics.

\bibitem[{Nivre et~al.(2018)Nivre, Abrams, Agi{\'c}, Ahrenberg, Antonsen,
  Aranzabe, Arutie, Asahara, Ateyah, Attia et~al.}]{nivre2018universal}
Joakim Nivre, Mitchell Abrams, {\v{Z}}eljko Agi{\'c}, Lars Ahrenberg, Lene
  Antonsen, Maria~Jesus Aranzabe, Gashaw Arutie, Masayuki Asahara, Luma Ateyah,
  Mohammed Attia, et~al. 2018.
\newblock Universal dependencies 2.2.

\bibitem[{Pires et~al.(2019)Pires, Schlinger, and
  Garrette}]{pires-etal-2019-multilingual}
Telmo Pires, Eva Schlinger, and Dan Garrette. 2019.
\newblock \href {https://doi.org/10.18653/v1/P19-1493} {How multilingual is
  multilingual {BERT}?}
\newblock In \emph{Proceedings of the 57th Annual Meeting of the Association
  for Computational Linguistics}, pages 4996--5001, Florence, Italy.
  Association for Computational Linguistics.

\bibitem[{Qi et~al.(2022)Qi, Wan, Du, and Chen}]{qi-etal-2022-enhancing}
Kunxun Qi, Hai Wan, Jianfeng Du, and Haolan Chen. 2022.
\newblock \href {https://doi.org/10.18653/v1/2022.acl-long.134} {Enhancing
  cross-lingual natural language inference by prompt-learning from
  cross-lingual templates}.
\newblock In \emph{Proceedings of the 60th Annual Meeting of the Association
  for Computational Linguistics (Volume 1: Long Papers)}, pages 1910--1923,
  Dublin, Ireland. Association for Computational Linguistics.

\bibitem[{Ruder et~al.(2021)Ruder, Constant, Botha, Siddhant, Firat, Fu, Liu,
  Hu, Garrette, Neubig, and Johnson}]{ruder-etal-2021-xtreme}
Sebastian Ruder, Noah Constant, Jan Botha, Aditya Siddhant, Orhan Firat, Jinlan
  Fu, Pengfei Liu, Junjie Hu, Dan Garrette, Graham Neubig, and Melvin Johnson.
  2021.
\newblock \href {https://doi.org/10.18653/v1/2021.emnlp-main.802}
  {{XTREME}-{R}: Towards more challenging and nuanced multilingual evaluation}.
\newblock In \emph{Proceedings of the 2021 Conference on Empirical Methods in
  Natural Language Processing}, pages 10215--10245, Online and Punta Cana,
  Dominican Republic. Association for Computational Linguistics.

\bibitem[{Shin et~al.(2020)Shin, Razeghi, IV, Wallace, and
  Singh}]{autoprompt:emnlp20}
Taylor Shin, Yasaman Razeghi, Robert L.~Logan IV, Eric Wallace, and Sameer
  Singh. 2020.
\newblock {AutoPrompt}: Eliciting knowledge from language models with
  automatically generated prompts.
\newblock In \emph{Empirical Methods in Natural Language Processing (EMNLP)}.

\bibitem[{van~der Maaten and Hinton(2008)}]{JMLR:v9:vandermaaten08a}
Laurens van~der Maaten and Geoffrey Hinton. 2008.
\newblock \href {http://jmlr.org/papers/v9/vandermaaten08a.html} {Visualizing
  data using t-sne}.
\newblock \emph{Journal of Machine Learning Research}, 9(86):2579--2605.

\bibitem[{Winata et~al.(2021)Winata, Madotto, Lin, Liu, Yosinski, and
  Fung}]{winata-etal-2021-language}
Genta~Indra Winata, Andrea Madotto, Zhaojiang Lin, Rosanne Liu, Jason Yosinski,
  and Pascale Fung. 2021.
\newblock \href {https://doi.org/10.18653/v1/2021.mrl-1.1} {Language models are
  few-shot multilingual learners}.
\newblock In \emph{Proceedings of the 1st Workshop on Multilingual
  Representation Learning}, pages 1--15, Punta Cana, Dominican Republic.
  Association for Computational Linguistics.

\bibitem[{Wolf et~al.(2020)Wolf, Debut, Sanh, Chaumond, Delangue, Moi, Cistac,
  Rault, Louf, Funtowicz, Davison, Shleifer, von Platen, Ma, Jernite, Plu, Xu,
  Le~Scao, Gugger, Drame, Lhoest, and Rush}]{wolf-etal-2020-transformers}
Thomas Wolf, Lysandre Debut, Victor Sanh, Julien Chaumond, Clement Delangue,
  Anthony Moi, Pierric Cistac, Tim Rault, Remi Louf, Morgan Funtowicz, Joe
  Davison, Sam Shleifer, Patrick von Platen, Clara Ma, Yacine Jernite, Julien
  Plu, Canwen Xu, Teven Le~Scao, Sylvain Gugger, Mariama Drame, Quentin Lhoest,
  and Alexander Rush. 2020.
\newblock \href {https://doi.org/10.18653/v1/2020.emnlp-demos.6} {Transformers:
  State-of-the-art natural language processing}.
\newblock In \emph{Proceedings of the 2020 Conference on Empirical Methods in
  Natural Language Processing: System Demonstrations}, pages 38--45, Online.
  Association for Computational Linguistics.

\bibitem[{Wu and Dredze(2019)}]{wu-dredze-2019-beto}
Shijie Wu and Mark Dredze. 2019.
\newblock \href {https://doi.org/10.18653/v1/D19-1077} {Beto, bentz, becas: The
  surprising cross-lingual effectiveness of {BERT}}.
\newblock In \emph{Proceedings of the 2019 Conference on Empirical Methods in
  Natural Language Processing and the 9th International Joint Conference on
  Natural Language Processing (EMNLP-IJCNLP)}, pages 833--844, Hong Kong,
  China. Association for Computational Linguistics.

\bibitem[{Yang et~al.(2019)Yang, Zhang, Tar, and
  Baldridge}]{yang-etal-2019-paws}
Yinfei Yang, Yuan Zhang, Chris Tar, and Jason Baldridge. 2019.
\newblock \href {https://doi.org/10.18653/v1/D19-1382} {{PAWS}-{X}: A
  cross-lingual adversarial dataset for paraphrase identification}.
\newblock In \emph{Proceedings of the 2019 Conference on Empirical Methods in
  Natural Language Processing and the 9th International Joint Conference on
  Natural Language Processing (EMNLP-IJCNLP)}, pages 3687--3692, Hong Kong,
  China. Association for Computational Linguistics.

\bibitem[{Zhang et~al.(2019)Zhang, Han, Liu, Jiang, Sun, and
  Liu}]{zhang-etal-2019-ernie}
Zhengyan Zhang, Xu~Han, Zhiyuan Liu, Xin Jiang, Maosong Sun, and Qun Liu. 2019.
\newblock \href {https://doi.org/10.18653/v1/P19-1139} {{ERNIE}: Enhanced
  language representation with informative entities}.
\newblock In \emph{Proceedings of the 57th Annual Meeting of the Association
  for Computational Linguistics}, pages 1441--1451, Florence, Italy.
  Association for Computational Linguistics.

\bibitem[{Zhao and Sch{\"u}tze(2021)}]{zhao-schutze-2021-discrete}
Mengjie Zhao and Hinrich Sch{\"u}tze. 2021.
\newblock \href {https://doi.org/10.18653/v1/2021.emnlp-main.672} {Discrete and
  soft prompting for multilingual models}.
\newblock In \emph{Proceedings of the 2021 Conference on Empirical Methods in
  Natural Language Processing}, pages 8547--8555, Online and Punta Cana,
  Dominican Republic. Association for Computational Linguistics.

\bibitem[{Zhong et~al.(2021)Zhong, Friedman, and
  Chen}]{zhong-etal-2021-factual}
Zexuan Zhong, Dan Friedman, and Danqi Chen. 2021.
\newblock \href {https://doi.org/10.18653/v1/2021.naacl-main.398} {Factual
  probing is [{MASK}]: Learning vs. learning to recall}.
\newblock In \emph{Proceedings of the 2021 Conference of the North American
  Chapter of the Association for Computational Linguistics: Human Language
  Technologies}, pages 5017--5033, Online. Association for Computational
  Linguistics.

\end{thebibliography}
\bibliographystyle{acl_natbib}

\appendix

\section{Appendix}
\label{sec:appendix}

\subsection{More Training Details}
For prompt tuning, we train with the Adam optimizer \citep{kingma2015adam} with no warmup step. Batch size is 32 for tasks, and with the exception of answering questions, which has a batch size of 8. Linear learning rate scheduler is used. We tune the learning rate in $\{5\mathrm{e}\!-\!2, 1\mathrm{e}\!-\!2, 5\mathrm{e}\!-\!3, 1\mathrm{e}\!-\!3, 5\mathrm{e}\!-\!4, 1\mathrm{e}\!-\!4\}$. We train all prompt tuning models for 30 epochs. Finally, tuned prompt length for MNLI is 32. It is 16 for the other tasks. We use A100s with 40G memory and all experiments can be done in few hours.

\begin{table*}[!h]
\resizebox{\textwidth}{!}{
\begin{tabular}{l|ccccccccccccccc|c}
\toprule
Method                                & en   & ar   & bg   & de   & el   & es   & fr   & hi   & ru   & sw   & th   & tr   & ur   & vi   & zh   & \textbf{avg}  \\
\midrule
\multicolumn{1}{l|}{}        &  88.2       & 77.4       & 82.3        & 82.6      &  81.1        &   83.7        &      82.0    &    75.2     &   79      &   71.0     & 76.7         &  77.5         & 71.4        &    79.1      & 78.6         &  79.1       \\
\multicolumn{1}{l|}{}         &   88.3       &  76.9       &      81.9    &   81.9       &  81.4         &     83.6     &    81.6      &   74.3      &  78.1       &  70.1         &   75.8       &  77.6         & 70.7         & 78.8         &   77.6       &  78.6         \\
\multicolumn{1}{c|}{\multirow{1}{*}{FT}} & 88.1 & 77.5 & 82.4 & 81.8 & 81.3 &  83.4 & 82.6 & 75.0 & 78.9 & 70.3 & 75.6 & 78.1 & 70.8& 78.5& 78.3 &78.8 \\
\multicolumn{1}{l|}{}         &   88.4     & 77.4        & 81.7        & 82.0       &   81.5       &     83.3     &  82.3        &     75.4    &  79.0       &  70.2      & 75.5       &    78.1     &  71.2     &   79.1    &  77.9        & 78.9       \\
\multicolumn{1}{l|}{}         &   88.2       & 77.6        &   82.5       &  81.7         &   80.9       &     83.2    & 81.9         &   75.1      &  78.2        &    69.5     &   76.5      &   77.6      &  71.0        & 78.8        &  78.6     & 78.8      \\
\midrule

\multicolumn{1}{l|}{}        &    88.5     & 78.3       &  82.8       &   82.2     &    82.5      &  84.2         &   83.0       &   76.1      &   80.4      &   71.0     &   77.6       & 79.2          &   72.5      &   80.0       &     78.3     &   79.8      \\
\multicolumn{1}{l|}{}         &    88.7      &   78.7      &  82.9        &  82.1        & 82.8          &  84.3        & 83.2         &      76.1   &      80.4   &   71.0        &  77.6        &    79.2       &   72.5       & 80.0         & 78.3         &      79.8     \\
\multicolumn{1}{c|}{\multirow{1}{*}{PT}} & 88.8 & 78.1 & 82.7 & 81.7 & 81.9 & 84.0 & 83.2 & 75.9 & 80.7 & 71.4 & 77.5 & 79.3 & 72.5& 79.4& 78.7 & 79.7\\
\multicolumn{1}{l|}{}         &  89.1       & 79.2        &  83.2       &    82.1    & 82.4         &  84.1        &  83.0        &     76.2    &   80.8      & 70.7       & 77.7       &    79.5     &  72.5     &     79.9  &   78.4       &  79.9      \\
\multicolumn{1}{l|}{}         &    89.0      &   78.7      &  83.2        &   82.2        &       82.8   &   84.3      &  83.4        &   76.2      &    80.8      &  71.3       &    77.9     &   79.2      & 72.5         & 80.3        &   78.2    & 80.0      \\
\bottomrule
\end{tabular}}
\caption{XNLI accuracy scores for each language with fine-tuning (FT) and prompt tuning (PT).}
\label{tab:xnli_results}
\end{table*}

\begin{table*}[!ht]
\centering
\resizebox{\columnwidth}{!}{
\begin{tabular}{l|ccccccc|c}
\toprule
Method                                                                           & en   & de   & es   & fr   & ja   & ko   & zh   & \textbf{avg}  \\
\midrule
\multicolumn{1}{l|}{}       &  95.6      & 90.8        &  81.4        &  91.3      &  82.7        &   81.8       &     84.5     &    88.3     \\
\multicolumn{1}{l|}{}       &     95.7   &   90.5      &    91.0      & 91.3       &    81.7      &    81.2      &  84.0        &    87.9     \\
\multicolumn{1}{c|}{\multirow{1}{*}{FT}}       &    95.4    &    89.4     &  90.8        &    90.9    & 80.5         &    80.6      &   84.0       &   87.4      \\
\multicolumn{1}{l|}{}       &  95.4      &    90.2     &  90.6        &   90.5     &      80.6    &    80.4      &       83.4   &      87.2   \\
\multicolumn{1}{l|}{}       &  94.7      &    91.0     &  91.4        &   92.1     &     82.4     &      93.2    &  84.2        &  88.6       \\
\midrule
\multicolumn{1}{l|}{}       &   96.2     &    92.3     &  91.4        &  92.1      &   81.3       &  83.2        & 84.8         &    88.8     \\
\multicolumn{1}{l|}{}       &   95.3     &   91.6      &   91.1       &   92.0     &   82.7       & 83.1         &  84.2        &     88.6    \\
\multicolumn{1}{c|}{\multirow{1}{*}{PT}}       &    95.4    &  90.9       &  91.4        &  91.8      &   82.1       &  82.8        &      84.7    &  88.4       \\
\multicolumn{1}{l|}{}       &  95.9      &  90.7       &  90.7        &   91.6     & 81.4         & 81.6         &   84.6       &      88.1   \\
\multicolumn{1}{l|}{}       &   95.6     &   91.6      &   90.5       &   91.7     &  82.2        &   81.7       &    83.0      &     88.0    \\
\bottomrule
\end{tabular}}
\caption{PAWS-X accuracy scores for each language with fine-tuning (FT) and prompt tuning (PT).}
\label{tab:paws-x-results}
\end{table*}

\begin{table*}[]
\resizebox{\textwidth}{!}{
\begin{tabular}{l|ccccccccccc|c}
\toprule
Method                                                                           & en          & es          & de          & el          & ru          & tr          & ar          & vi          & th          & zh          & hi          & \textbf{avg}         \\
\midrule
 \multicolumn{1}{l|}{} & 75.2 / 87.2 & 61.2 / 80.7 & 62.7 / 82.5 & 60.2 / 78.7 & 63.5 / 80.1 & 57.8 / 74.3 & 58.6 / 75.5 & 59.9 / 79.4 & 59.9 / 73.2 & 58.7 / 68.5 & 56.6 / 74.7 & 61.3 / 77.7 \\
\multicolumn{1}{l|}{} & 75.0 / 86.8 & 61.6 / 79.8 & 61.9 / 80.0 & 59.6 / 78.6 & 62.7 / 79.6 & 57.6 / 73.3 & 56.6 / 74.4 & 57.8 / 78.6 & 61.3 / 72.4 & 60 / 67.5   & 58.2 / 74.6 & 61.1 / 76.9 \\
\multicolumn{1}{l|}{\multirow{1}{*}{PT}} & 75.5 / 87.0   & 64.0 / 81.3 & 64.8 / 80.9 & 62.4 / 80.0 & 63.8 / 80.1 & 57.7 / 73.8 & 55.9 / 72.8 & 60.2 / 79.5 & 62.3 / 73.4 & 61.4 / 69.6 & 59.8 / 76.1 & 62.5 / 77.7 \\
\multicolumn{1}{l|}{} & 75.8 / 87.0   & 63.0 / 81.4 & 62.4 / 79.4 & 62.1 / 79.9 & 62.9 / 79.8 & 56.9 / 73.6 & 57.4 / 74.6 & 59.6 / 78.5 & 62.8 / 74.7 & 60.5 / 70.5 & 57.5 / 74.2 & 61.9 / 77.6 \\
\multicolumn{1}{l|}{} & 76.0 / 87.4 & 62.8 / 80.8 & 65.0 / 80.2 & 61.2 / 78.3 & 63.1 / 79.5 & 56.3 / 72.3 & 57.3 / 73.9 & 57.6 / 77.5 & 62.9 / 71.6 & 61.0 / 68.7 & 58.4 / 74.2 & 62.0 / 76.8   \\
\midrule
 \multicolumn{1}{l|}{} & 77.2 / 88.4 & 65.1 / 83.1 & 64.8 / 81.4 & 63.7 / 81.2 & 58.7 / 80.2 & 58.7 / 74.6 & 60.3 / 77.0 & 61.4 / 80.6 & 66.4 / 74.7 & 60.3 / 68.6 & 61.8 / 78.1 & 63.5 / 78.9 \\
\multicolumn{1}{l|}{} & 77.4 / 88.5 & 64.4 / 82.3 & 64.8 / 81.2 & 63.5 / 80.8 & 64.7 / 80.7 & 58.3 / 74.1 & 60.3 / 76.8 & 61.0 / 80.3 & 66.6 / 75.0 & 61.7 / 70.2 & 61.5 / 77.5 & 64.0 / 78.9 \\
 \multicolumn{1}{l|}{\multirow{1}{*}{PT}}  & 77.4 / 88.6 & 65.4 / 83.4 & 64.5 / 80.9 & 64.0 / 81.2 & 64.1 / 80.7 & 58.7 / 74.9 & 59.8 / 76.5 & 62.1 / 81.4 & 66.6 / 75.4 & 61.3 / 69.9 & 62.8 / 77.8 & 64.2 / 79.2 \\
\multicolumn{1}{l|}{} & 77.1 / 88.5 & 64.7 / 82.9 & 63.9 / 80.7 & 62.7 / 80.5 & 64.7 / 80.4 & 59.2 / 74.6 & 59.7 / 76.3 & 60.8 / 80.7 & 66.6 / 74.6 & 61.0 / 69.1 & 61.6 / 77.6 & 64.7 / 78.7 \\
\multicolumn{1}{l|}{} & 77.9 / 88.7 & 65.0 / 83.0 & 64.2 / 81.2 & 63.4 / 80.2 & 64.8 / 80.9 & 58.2 / 75.4 & 60.2 / 77.0 & 62.9 / 81.3 & 67.3 / 75.9 & 60.7 / 69.8 & 61.1 / 77.9 & 64.2 / 79.2 \\
\bottomrule
\end{tabular}}
\caption{XQuAD results (EM / F1) for each language with fine-tuning (FT) and prompt tuning (PT).}
\label{tab:xquad-results}
\end{table*}

\begin{table*}[]
\centering
\resizebox{\textwidth}{!}{%
\begin{tabular}{l c c c c c c c c c c}
\toprule
Method & en & ar & bn & fi & id & ko & ru & sw & te & avg \\ \midrule
\multicolumn{1}{l|}{} & 60.5 / 74.2 & 51.5 / 71.5 & 50.4 / 68.6 & 51.0 / 67.6 & 62.5 / 78.6 & 49.6 / 60.9 & 45.3 / 67.7 & 44.7 / 65.7 & 56.7 / 75.7 & 46.2 / 70.0 \\
\multicolumn{1}{l|}{} & 57.7 / 71.8 & 51.5 / 71.0 & 50.4 / 70.4 & 53.5 / 70.3 & 61.4 / 77.1 & 53.6 / 64.5 & 47.8 / 68.6 & 50.3 / 70.3 & 57.0 / 75.7 & 53.7 / 71.1 \\
\multicolumn{1}{l|}{\multirow{1}{*}{FT}} & 57.7 / 73.2 & 51.9 / 72.5 & 48.7 / 66.4 & 53.6 / 69.7 & 59.8 / 77.0 & 50.7 / 59.6 & 50.7 / 68.0 & 49.1 / 69.1 & 58.0 / 77.8 & 53.4 / 70.4 \\
\multicolumn{1}{l|}{} & 58.6 / 71.7 & 53.0 / 71.6 & 46.0 / 62.5 & 53.7 / 68.4 & 59.8 / 75.7 & 52.5 / 63.0 & 40.4 / 65.2 & 48.9 / 69.2 & 58.4 / 76.3 & 52.4 / 69.3 \\
\multicolumn{1}{l|}{} & 59.8 / 72.3 & 48.3 / 70.6 & 52.2 / 68.6 & 49.7 / 67.5 & 60.4 / 77.7 & 55.1 / 65.5 & 38.9 / 65.2 & 45.4 / 66.6 & 56.8 / 75.4 & 51.8 / 69.9 \\

 \midrule
\multicolumn{1}{l|}{} & 61.8 / 75.0 & 53.7 / 72.3 & 48.7 / 67.0 & 58.2 / 73.0 & 63.0 / 77.9 & 52.9 / 63.6 & 50.2 / 70.0 & 47.5 / 68.5 & 57.5 / 75.3 & 54.8 / 71.5 \\
\multicolumn{1}{l|}{} & 60.7 / 74.0 & 53.1/72.2 & 45.1 / 64.5 & 55.9 / 71.8 & 63.5 / 78.3 & 51.8 / 61.9 & 52.3 / 71.0 & 48.9 / 68.9 & 58.4 / 76.2 & 54.4 / 71.0 \\
\multicolumn{1}{l|}{\multirow{1}{*}{PT}} & 60.2 / 73.6 & 54.8 / 73.9 & 52.2 / 70.0 & 56.6 / 71.4 & 64.8 / 78.7 & 52.5 / 62.3 & 53.1 / 71.4 & 51.1 / 70.7 & 61.6 / 79.1 & 56.3 / 72.3 \\
\multicolumn{1}{l|}{} & 62.0 / 75.3 & 53.6 / 73.0 & 46.0 / 64.9 & 57.3 / 71.3 & 63.7 / 78.6 & 53.3 / 62.0 & 52.7 / 71.8 & 48.1 / 69.0 & 58.7 / 75.5 & 55.0 / 71.3 \\
\multicolumn{1}{l|}{} & 61.4 / 74.5 & 54.9 / 72.8 & 46.9 / 66.3 & 56.8 / 71.4 & 63.2 / 77.6 & 54.3 / 63.0 & 53.1 / 71.1 & 47.9 / 68.4 & 58.6 / 76.4 & 55.2 / 71.3 \\
\bottomrule
\end{tabular}%
}
\caption{TyDiQA-GoldP results (EM / F1) for each language with fine-tuning (FT) and prompt tuning (PT).}
\label{tab:tydiqa_results}
\end{table*}

\end{document}